# Decision support system for forest fire management using Ontology with Big Data and LLMs

Ritesh Chandra*, Shashi Shekhar Kumar, Rushil Patra, and Sonali Agarwal
Indian Institute of Information Technology Allahabad, India
*rsi2022001@iiita.ac.in, rsi2020502@iiita.ac.in, iib2020005@iiita.ac.in, sonali@iiita.ac.in,*

**Abstract**

Forests are crucial for ecological balance, but wildfires are a major cause of forest loss, posing significant risks. Fire weather indices, which assess wildfire risk and predict resource demands, are vital. Semantic sensor networks, gaining popularity in fields like healthcare and environmental monitoring, are increasingly gathering climatic data like wind speed, temperature, and humidity through these networks. However, processing these data streams to determine fire weather indices presents challenges, underscoring the growing importance of effective forest fire detection. This paper discusses using Apache Spark for early forest fire detection, enhancing fire risk prediction with meteorological and geographical data. Building on our previous development of Semantic Sensor Network (SSN) ontologies and Semantic Web Rules Language (SWRL) for managing forest fires in Monesterial Natural Park, we expanded SWRL to improve a Decision Support System (DSS) using a Large Language Models (LLMs) and Spark framework. We implemented real-time alerts with Spark streaming, tailored to various fire scenarios, and validated our approach using ontology metrics, query-based evaluations, LLMs score precision, F1 score, and recall measures.

## 1. Introduction

Forest fires constitute a crucial worldwide undertaking[1], posing giant threats to ecological integrity and monetary balance. Every year, these fires consume tens of thousands of hectares of forests globally, resulting in the loss of vital habitats, a reduction in biodiversity, and the discharge of substantial amounts of carbon emissions into the atmosphere, exacerbating the effects of climate change. While herbal phenomena like lightning can provoke forest fires in particular situations, the major causes are human activities, consisting of agricultural burning, land clearing, and accidental ignition.

India is particularly vulnerable to forest fires[2] due to a variety of factors, including significant human encroachment near forested regions, unsustainable land-use practices, and inadequate fire management strategies. Research indicates that a significant number of fires in India originate from human activity rather than natural causes. These fires now devastate woodland ecosystems, endangered wildlife, and directly harm human communities living in or near woodland areas. Moreover, these fires’ smoke and particulate matter contribute considerably to air pollutant levels, compromising public health and environmental excellence.

---

1 https://www.c2es.org/content/wildfires-and-climate-change/#:~:text=Climate%20change%20enhances%20the%20drying,climate%20create%20warmer%2C%20drier%20conditions.

2 https://fsi.nic.in/index.php

The escalating influences of climate change exacerbate the problem by extending the period and intensifying the severity of heart seasons globally. Consequently, there is a pressing need to develop and implement more powerful and localized fire management and prevention strategies. Most forest management companies rely closely on fireplace threat score systems to assess health hazards and allocate firefighting sources. However, several current worldwide fireplace chance score systems, like the Canadian Fire Weather Index (CFWI) [1] and the McArthur Forest Fire Danger Index [2], rely on weather and ecological data from regions with unique climatic and ecological characteristics. As a result, these systems might not reflect the unique and diverse conditions universal in Indian forests.

Spark is a unified analytics engine that efficiently performs several operations on big data. It can easily integrate with LLMs to create an efficient real-time DSS. This integration is beneficial when working with vast amounts of data to interpret them and make automated reasoning more accurate [3]. Working with Spark and LLMs enables us to make accurate decisions based on prespecified rules and environments for real-time operations. LLMs can perform automated reasoning on Spark-processed data and integrate the capability to improve the logic and accuracy of rule-based queries, particularly for complex decision-making tasks.

In order to establish a unified data model within the information space, ontologies streamline several critical aspects. (1) They give detailed information on a wide range of data sources and infrastructure services, including insights specific to the domain; (2) They create a common language that makes it easier for the infrastructure to interact with outside sources that use different methods; and (3) They make it easier to find data, access it, and combine it. These ontologies cater to a wide range of knowledge representation requirements: (1) The SSN ontology specifically models sensor networks and captures data concerning distinct features of interest. In this framework, DOLCE+DnS UltraLite serves as the upper ontology; (2) it showcases the resources provided by the infrastructure and the accessible datasets.

This research work addresses several challenges:

1. Real-time data processing with Spark enables handling large volumes of sensor data to support quick decisions. Integration with LLMs ensures that varying data volumes do not affect the model's performance or uptime.

2. It identifies suitable data sources based on their content, which includes features, observations, sensors, and dataset-defined scope.

3. It tackles the heterogeneity of data sources, including different data models, interfaces, and mobility issues.

4. Users participate in data collection based on domain concepts, synthesizing information from various sources to improve problem understanding.

The proposed approach utilizes semantic technologies in decision support systems for forest fires to manage diverse data sources and facilitate data correlation. User-level applications can create ontology-based queries and then transform them into executable queries for data sources. Spark streaming with LLMs allows the system to interact with textual data in real time and make effective decisions based on a predefined set of rules and queries.

We designed the system to achieve the following objectives and meet specific user needs:

1. Develop a SWRL for a rule-based DSS that can be represented using classes, properties, and individuals.
2. To present a real-time LLMs-based alert generation model.
3. • To establish an index, formulated with meteorological experts' guidance, to showcase fire weather threat ratings, using Ontology Web Language (OWL)
4. To effectively manage and process large-scale sensor data.
5. We transform raw data into Resource Description Framework (RDF) triples to enhance data quality.

The remaining sections of this research work are structured as follows: Section II presents the background and related work to identify the research gap. Section III expounds on the suggested research methodology. Section IV presents the experimental works and their results. Section V discusses the conclusions and recommendations for future work. Table 1 shows the abbreviations used in this manuscript.

Table 1: List of abbreviations

| **Abbreviations** | **Name** |
|---|---|
| SSN | Semantic Sensor Network |
| DSS | Decision Support System |
| LLMs | Large Language Models |
| OWL | Ontology Web Language |
| RDF | Resource Description Framework |
| CFWI | Canadian Fire Weather Index |
| BUI | Buildup Index |
| FFMC | Fine Fuel Moisture Code |
| SWRL | Semantic Web Rule Language |
| FWI | Fire Weather Index |
| DMC | Duff Moisture Code |
| DC | Drought Code |
| ISI | Initial Spread Index |
| SVM | Support Vector Machine |

## 2. Literature review

In this section, we will discuss the related works done in recent years and identify the research gaps that need further exploration using the underlying methodology shown in table 2. Recent advancements in forest fire detection have seen a variety of innovative approaches leveraging deep learning, IoT, and wireless technologies.

Abdusalomov et al. [4] created an automated method using an improved Detectron2 platform. This method was able to find fires with an amazing 99.3% accuracy in a variety of conditions, both during the day and at night. This method utilized a custom dataset and significantly improved fire detection accuracy.

Burak et al. [5] suggested a hierarchical forest fire detection system that uses a mix of multimedia and scalar sensors in IoT systems. This system was 98.28% accurate in detecting fires and used 29.94% less energy.

Avazov et al. [6] introduced a fire detection method tailored for smart cities, leveraging an enhanced YOLOv4-based convolutional neural network capable of triggering an alarm within eight seconds of a fire outbreak with high speed and accuracy.

Karim et al. [7] used mid-point K-means clustering with Wireless Sensor Networks (WSNs) to accurately predict 98% of the time where there would be a lot of activity in the forest while still making sure that sending data used as little energy as possible.

Bharany et al. [8] studied how to use Flying Ad Hoc Networks (FANETs) to find fires. They used the EE-SS clustering algorithm, which made UAVs last longer and worked better than other cutting-edge algorithms. These studies show that advanced technologies like deep learning, IoT, and WSNs are being used more and more to make forest fire detection systems more accurate, faster, and more efficient.

Table 2: Summary of existing works

| **Author-Years** | **Objectives** | **Proposed Methodology** | **Results** | **Tools and Languages** |
|---|---|---|---|---|
| Ginkal et al 2024[9] | Cloud-based wireless sensor network technology for forest fire detection. | WSN, Computer Vision, UAV, YOLO, IoT, MODIS | Real-time monitoring improved efficiency over traditional methods | WSN, Computer Vision, UAV, YOLO, IoT, MODIS |
| Ibraheem et al.2024[10] | The early detection system employs the Intermediate Fusion VGG16 model and ECP-LEACH. | Intermediate Fusion VGG16, ECP-LEACH | 99.86% detection accuracy, regulated energy consumption | Python, Intermediate Fusion VGG16, ECP-LEACH |
| Huiyi et al. 2024[11] | Identifying causes of forest fire events | The Jenk-DBSCAN model, MODIS MCD64A1 product, RF, LR, and SVM | The accuracy in identifying fire causes is high, with meteorological variables being the primary factors. | enk-DBSCAN, MODIS, RF, LR, SVM |
| Cumbane et al. 2019 [12] | Big data framework for disaster response application. | Review various frameworks and how efficiently it can be used to solve natural disasters. | Evaluating each framework on various parameters such as scalability, responsiveness, | Apache Spark, Kafka, and Hadoop. |
| | | | data processing, etc. | |
| Athanasis et al. 2019 [13] | Big Data Analysis for Wildfire Prevention and Management | Role of Geospatial big data for wildfire management. | contribute to cutting-edge knowledge sharing between wildfire conflict operation centers and field fire fighting units. | Big data clusters. |
| Rajasekaran et al. 2015[14] | Big data analysis for forest fire management. | Role of framework to real time alert generation. | Thirty minutes before forest fire prediction to reduce the casualties. | Apache Hadoop. |
| Martins et al. 2022 [15] | To develop a forest fire susceptibility mapping approach using machine learning in Southeast China | Employed space-humidity-constrained (SHC-based) sampling method and dynamic factor extraction method based on DTW and precipitation thresholds (Thcp) | RF model demonstrated the best performance with accuracy enhancements of over 8% | Random Forest (RF), Support Vector Machine (SVM), multi-layer perceptron (MLP) |
| Singh et al. 2024 [16] | To examine forest fire trends, patterns, and spatiotemporal distribution in tropical seasonal forests in Odisha State, India | Utilized MODIS imagery, Kernel density tools, and machine learning algorithms (SVM and RF) | Random Forest achieved over 94% precision in identifying forest fire susceptibility zones | SVM, Random Forest RF, MODIS imagery |
| Zhao et al.2024 [17] | To map forest fire sensitivity in Hunan Province using machine learning models | Constructed four machine learning models: RF, SVM, XGboost, and BRT | FR-BRT model performed the best with an AUC of 0.97 | RF (SVM), eXtreme Gradient Boosting (XGboost), Boosted |
| | | | | Regression Tree (BRT) |
| Mishra et al. 2024[18] | To develop a dual-model deep learning approach for super real-time wildfire spread forecasting | Utilized U-Net for prediction and ConvLSTM layers for real-time refinement | Over 90% agreement between AI forecasts and numerical simulations | U-Net, ConvLSTM, deep learning |
| Zhang et al. 2023 [19] | To predict post-fire burn severity in California using a novel machine-learning model | Employed the Super Learner (SL) algorithm accounting for spatial autocorrelation | SL model predicted burn severity with reasonable classification accuracy | SL, algorithm, Vecchia's Gaussian approximation |
| Li et al. 2024 [20] | To identify and prioritize forest fire-prone areas in Iran's Firouzabad region | Utilized GIS-based Bayesian and Random Forest methodologies | RF outperformed the Bayesian model with an AUC score of 0.876 | RF, Bayesian model, GIS |
| Simafranca et al. 2024[21] | It introduces an updated version of the ONTO-SAFE ontology with improved semantic capabilities, effectively capturing a range of forest fire scenarios. | Used Semantic Web technologies for developing ontology. | Based on metrics count got the result of Average population 70.089552 Class richness 0.044776 | SPARQL, RDF4j |
| Noroozi, et . 2024 [22] | Fine-tuned the LLMs for ontology construction due to limited domain expertise. | Utilizes prompt engineering. | Three ontology learning tasks were evaluated using fine-tuned LLMs. | LLaMA-7B,BLOOM-3B |

The literature review emphasizes the necessity of an effective model that integrates Spark streaming, LLMs, and ontology-based methods. This integrated approach aims to lower costs and achieve greater accuracy than current systems such as satellite-based, optical camera-based, and human-based forest fire observation systems. While various ontology-based decision support systems for forest fires exist, such as the BACAREX ontology, Geo-ontology, and FWI ontology, they primarily concentrate on observation, operational tracking, and FWI calculation. None of the reviewed studies address an integrated approach that merges LLMs and the Spark framework within a single ontology.

This research aims to create a framework within the SSN ontology that allows for the ability to search observation systems using SPARQL and improves interactivity for both computer programs and human users. By improving the understanding and application of forest fire and surveillance data, this all-encompassing framework seeks to increase the information's comprehensiveness. It uses Spark streaming, with the help of LLMs, to strengthen the framework and enable quick and efficient decision-making. The structure also illustrates the key components for controlling and monitoring forest fires.

## 3. Methodology

This section elaborates on the architecture and the corresponding work of the proposed research.

### 3.1 Architecture of the proposed model

Figure 1 illustrates the comprehensive working model, showcasing the proposed combination of Spark streaming, ontology, and LLMs. The application of SPARQL queries to both LLMs and ontology facilitates decision-making for forest fire management, based on the results obtained from this integration.

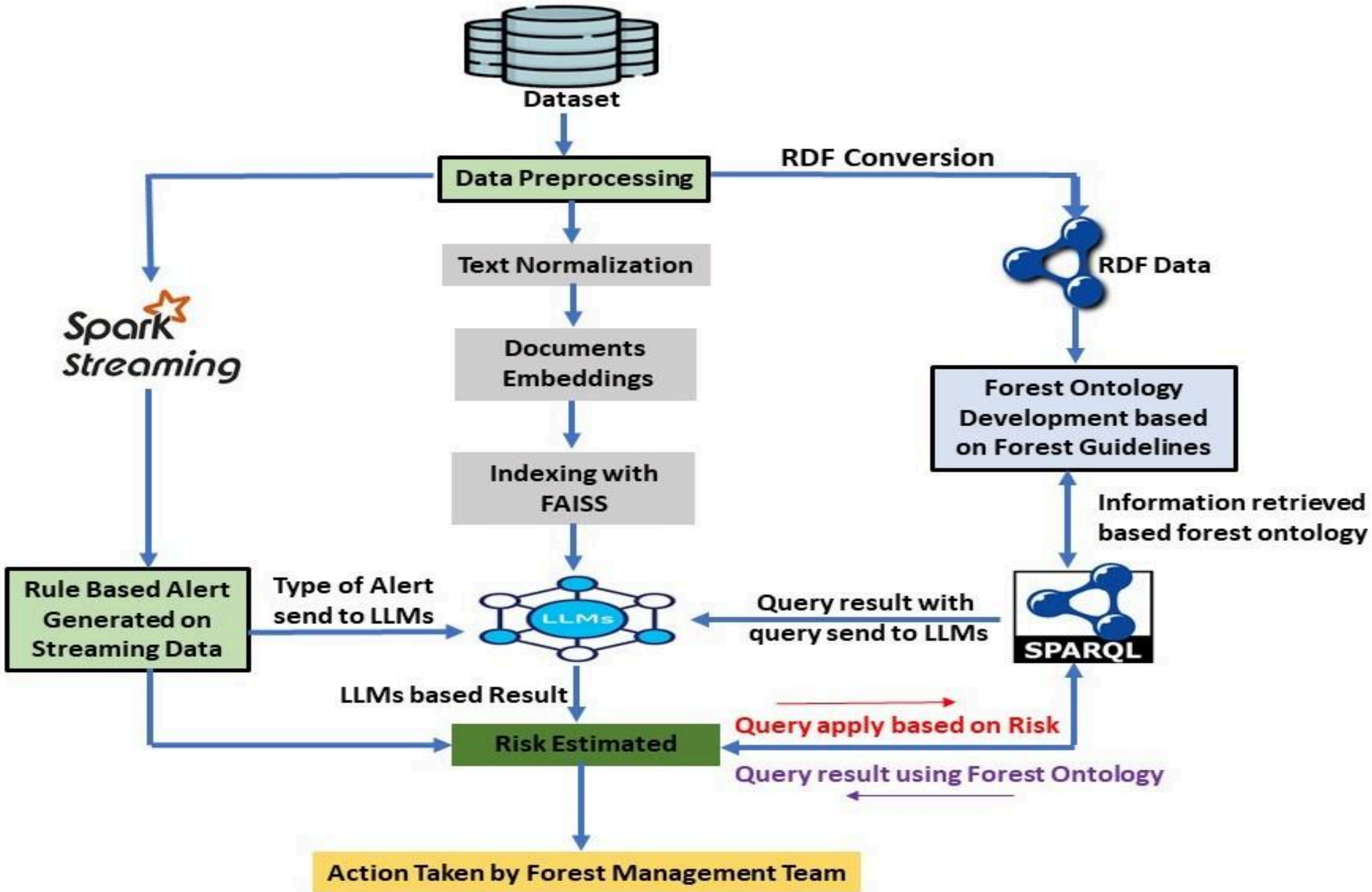

Figure 1: Framework view of the complete working model

### 3.2. Dataset description

The dataset encompasses a detailed array of spatial, temporal, and meteorological data crucial for analyzing and predicting forest fire behavior in Montesinho Park[3], Portugal. It has exact coordinates in space to find high-risk areas, information about time that shows how fires happen at different times of the year and week, and weather variables like temperature, humidity, wind speed, and rainfall that directly affect how fires spread. Furthermore, the dataset features a target variable, the burned area, which quantifies the extent of forest damage, providing valuable insights into fire severity. This variable, prone to skewness towards smaller values, is recommended to be transformed logarithmically for more accurate modeling. Overall, this dataset is instrumental for devising effective forest fire management and intervention strategies.

[3] https://amontesinho.pt/en/the-region/montesinho-natu-ral-park/

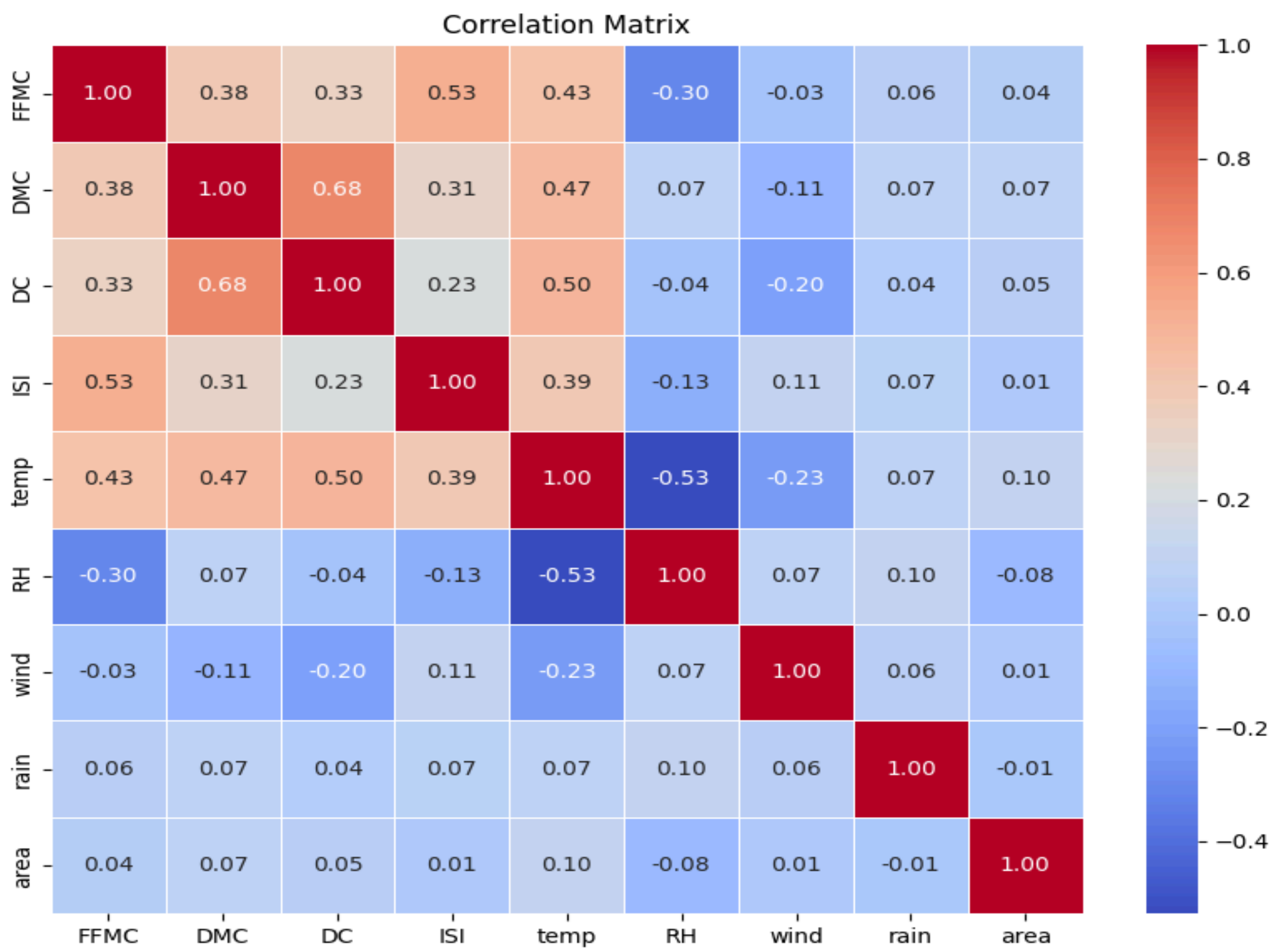

Figure 2: Correlation matrix of the dataset

Figure 2 matrix reveals a strong correlation between temperature, ISI, DC, DMC, and FFMC. These features also exhibit a positive correlation with the burned area, while RH, wind, and rain exhibit a slight positive correlation and a negative correlation with the spread area of the forest fire. This means that factors like RH, wind, and rain reduce the area of the forest fire, and factors, like temperature, ISI, DC, DMC, and FFMC combined can cause more fire spread.

### 3.2.1 Data Preprocessing

The data gathering and setup process starts with data streaming using PySpark and importing necessary libraries, followed by creating a stream context to convert real-time data into a continuous data frame stream for analysis. To clean up the data, we transform the skewed "area" target variable logarithmically to normalize its distribution, enabling its use in more models. Z-score normalization equalizes numerical features like FFMC, DMC, DC, ISI, temperature, RH, wind, and rain, making sure that there is no mean and no variance. This makes the model more stable and speeds up convergence. One-hot encoding turns categorical variables like "month" and "day" into binary columns, which makes it easier to use them in models that need numerical inputs.

To eliminate low-correlated or redundant features, we conduct feature selection through correlation analysis, thereby improving the model's efficiency. Statistical methods like z-scores or the interquartile range find and get rid of outliers by figuring out how they affect the model's performance and making changes as needed. For the "area" variable's unbalanced data, techniques like oversampling and undersampling even out the class distribution, make it easier for the model to find patterns.

### 3.2.2 RDF Conversion

RDF, developed in partnership with the World Wide Web Consortium (W3C), standardizes the description of online resources and facilitates seamless data transmission and integration. RDF makes clear connections between data elements by expressing them as subject-predicate-object triples. This lets you manage data in a flexible and structured way. One of RDF's key strengths is its ability to separate data from its structure, allowing for the integration of diverse data sources under a unified schema. In this research, we employ Python to transform a CSV dataset into RDF format, thereby facilitating data querying and linking across various sources. The CSV format, while simple to work with for structured data, lacks the semantic richness needed for more complex data modeling. Figure **3** shows the conversion into RDF data from CSV sources.

By converting the CSV data to RDF, we can enrich the dataset with metadata, making it more interoperable and accessible for linked data applications. Depending on the use case, we can serialize the RDF output from this conversion process in multiple formats like Turtle, XML, or JSON-LD.

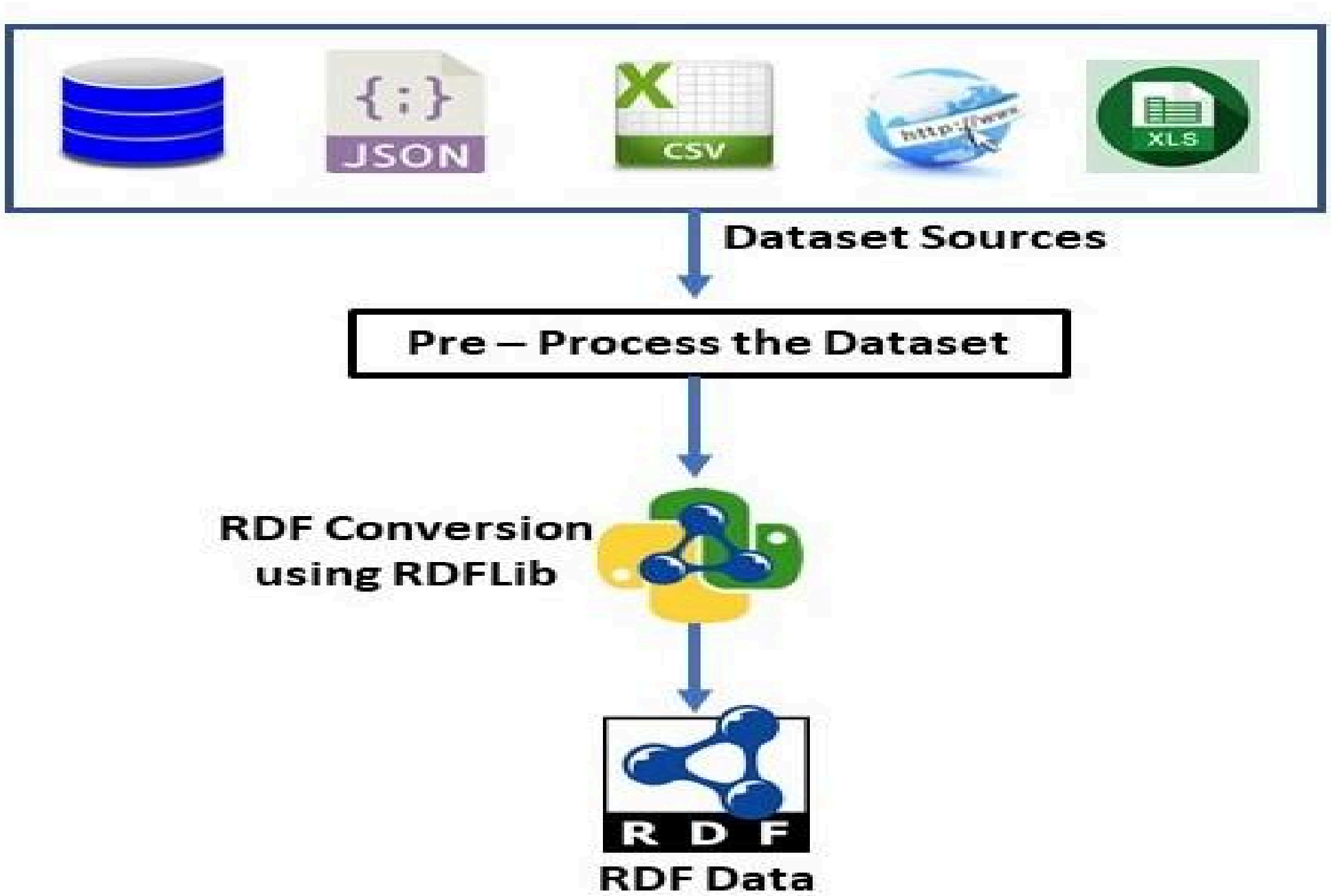

Figure 3: Converting datasets to RDF

### 3.3 Ontology Development

For the development of the forest ontology, we utilized the upper-level SSN ontology, which has a standardized forest ontology. The SSN ontology was developed by the Semantic Sensor Networks Incubator Group of the W3C. In our previous research, we based our ontology development on documents related to Canadian forest fires. However, in this current study, we have refined the forest ontology by incorporating documentation from other countries, like India, to increase its accuracy and to encompass a broader scope of knowledge regarding forest vegetation.

We have enhanced the ontology by incorporating a diverse array of policies, plans, and forest fire prevention rules, together with coordination strategies across agencies, as detailed in forest documentation from various countries. This improvement broadens the ontology's scope and applicability to forest fire management. Additionally, we have integrated numerous SWRL rules into the forest ontology, drawing on documentation from different nations to ensure a more robust and versatile framework.

#### 3.3.1 SWRL Rules Development

The SWRL[4] merges the OWL with the Rule Markup Language, offering a high-level abstract structure for rules similar to Horn clauses. This model's rules are formulated using OWL concepts such as classes, properties, and individuals. In our previous work, we incorporated 42 rules; however, in this current study, we have expanded the number to 110. Additionally, we have developed more complex rules based on our existing findings, details of which are presented in the following sections. These rule-based transformations are applied to streaming data.

Previous work established rules that depend on specific columns such as FFM, DMC, DC, ISI, rain, and wind. Various classes for each column value (e.g., extremely easy, difficult, fast) have been categorized using predefined thresholds. In this extended version, we enhance the work by covering all different scenarios, such as human resource management in emergencies and determining which type of firefighting equipment is needed at which time. This makes the model concerning forest management more complete. Some rules are shown in Tables 3, 4, and 5. We have accessed several documents[5] to develop these rules[6].

Table 3: SWRL rules for the elaboration of a fire management plan

| SWRL rules | Explanation |
| --- | --- |

[4] https://www.w3.org/Submission/SWRL

[5] https://www.rfs.nsw.gov.au/__data/assets/pdf_file/0013/71401/Training-Information-Booklet.pdf

[6] https://www.forest fire training guidelines

| | |
|---|---|
| PreventiveAction(?action)^hasScenario(?action,?scenario)^hasIgnitionRisk(?scenario, ?risk)^swrlb:lessThanOrEqual(?risk,0.5)->reduceIgnitionRisk(?action) | When the ignition risk of a scenario drops below a specified threshold (e.g., 0.5), indicating low risk, a rule is triggered. This rule recommends a preventive action to lower the ignition risk further. |
| PreventiveAction(?action)^hasScenario(?action, ?scenario)^hasBurnedArea(?scenario,?area)^swrlb:lessThanOrEqual(?area,1000)->limitBurnedArea(?action) | This rule suggests a preventive action to limit the burned area if the projected burned area for a scenario is below a certain threshold (e.g., 1000 hectares). |
| PreventiveAction(?action)^hasScenario(?action,?scenario)^hasDamage(?scenario, ?damage)^swrlb:lessThanOrEqual(?damage,20)->limitDamage(?action) | If the estimated damage for a scenario is below a specified threshold (e.g., 20%), this rule recommends an action to limit the damage caused by fire. |
| PreventiveAction(?action)^infrastructureInstallation(?action)->installInfrastructure(?action) | This rule suggests installing preventive infrastructure such as lookout towers, tracks, water points, or fuel breaks. |
| PreventiveAction(?action)^populationSensitization(?action) -> sensitizePopulation(?action) | This rule recommends sensitizing the population and involving them in preventive activities in the forest. |
| PreventiveAction(?action)^forestStandIntervention(?action)-> nterveneForestStand(?action) | This rule suggests interventions in forest stands, such as undergrowth clearing and other forest operations. |

Table 4: SWRL rules for human resources management in fire suppression

| **SWRL rules** | **Explanation** |
|---|---|
| SpecializedPersonnel(?person)^hasTraining(?person,"Firefighting")->deploySpecializedPersonnel(?person) | This rule suggests deploying personnel with specific firefighting training to handle forest fire suppression operations. |
| NotSpecializedPersonnel(?person)^hasInvolvement(?person,"Limited")->limitInvolvement(?person) | This rule recommends limiting the involvement of personnel without specialized firefighting training due to safety concerns and the potential for ineffective firefighting. |
| Firefighter(?person)^hasIndividualProtectiveEquipment(?person, "Protective Gear") -> ensure safety(?person) | Under this rule, firefighters are equipped with appropriate protective gear to ensure their safety during firefighting activities. |
| AdministrativeAuthority(?authority)^hasResponsibilityModel(?authority,"IntegratedForester")->delegateResponsibility(?authority,"ForestManagementServices") | This rule specifies the delegation of responsibility for forest fire suppression to forest management services under the "Integrated Forester" model. |
| FireSuppressionService(?service)^hasClearOrganization(?service,true)->ensureCoordination(?service) | This rule emphasizes the importance of clear organization within the fire suppression service to ensure effective coordination during firefighting operations. |
| Zone(?zone) ^ hasRiskLevel(?zone, "High") ->deployOptimalDensity(?zone,"OneBrigadePer5000Ha") | This rule suggests deploying an optimal density of firefighting personnel, such as one brigade per 5000 hectares, in zones with high fire risk. |

Table 5: SWRL rule for firefighting equipment selection criteria

| SWRL rules | Explanation |
|---|---|
| FireSuppressionOperation(?operation)^hasFireType(?operation,"Surface")->deployTerrestrialEquipment(?operation,"StandardEquipment") | This rule suggests deploying terrestrial equipment, specifically standard equipment like shovels, hoe-rakes, and fire beaters, when the fire type is surface fire. This equipment is effective for initial attacks and low-intensity fires. |
| FireSuppressionOperation(?operation)^hasDevelopmentPhase(?operation,"LargeUncontrolledFire") -> useInitialAttackVehicles(?operation) | When facing large, uncontrolled fires, this rule recommends utilizing initial attack vehicles equipped with water tanks. These vehicles provide immediate response capabilities and can suppress or slow down fires until reinforcements arrive. |
| WaterTanker(?vehicle)^hasWaterCapacity(?vehicle, ?capacity) ^ swrlb:lessThan(?capacity, 5000) -> deployMultipleVehicles(?vehicle) | For water tankers with capacities less than 5000 liters, it suggests deploying multiple vehicles instead of a single large one. This ensures flexibility and mobility while maintaining sufficient water supply for firefighting operations. |
| FireSuppressionOperation(?operation)^requiresIndirectIntervention(?operation,true)->deployBulldozers(?operation) | In cases where indirect intervention occurs, such as building fire lines, this rule recommends deploying bulldozers. These machines can clear vegetation to create fire breaks, limiting the spread of the fire. |
| FireSuppressionOperation(?operation)^hasEnvironmentalConditions(?operation,"DifficultAccess") -> useHelicopters(?operation) | In areas with difficult access, such as rugged terrain, this rule suggests using helicopters for aerial firefighting operations. Helicopters are more maneuverable and can access remote areas where fixed-wing aircraft might face challenges. |
| FireSuppressionOperation(?operation)^hasFireBehavior(?operation,"RapidSpread")->useShortTermRetardants(?operation) | When dealing with fires exhibiting rapid spread, this rule recommends using short-term retardants as chemical additives. These retardants increase water retention in vegetation, helping to slow down the fire's advance. |

### 3.4 Spark Streaming

Spark streaming is the core of the Apache Spark API, which provides scalable solutions for high throughput and fault tolerance for processing live data streams. It allows both real-time processing and batch processing to maintain a low-latency approach for various applications. It allows replication of data across various nodes, which makes Spark streaming more robust for handling data in case of failure. Being part of the Apache Spark platform, Spark Streaming integrates seamlessly with other Spark components, such as Spark SQL and MLlib (for machine learning).  This integration enables complex workflows that include streaming, batch processing, and interactive queries. Figure 4 shows Spark Streaming processing data in batches.

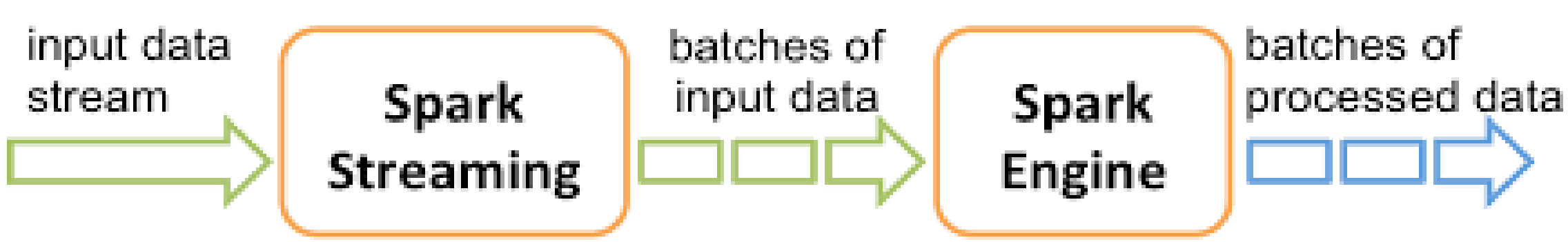

Figure 4: Stream processing through spark framework.

#### 3.4.1 Spark Streaming based alerts

To detect events and generate alerts on streaming data, we implement rules based on the FWI. The development of these rules incorporates parameters such as the Fine Fuel Moisture Code (FFMC), Duff Moisture Code (DMC), Drought Code (DC), Initial Spread Index (ISI),

Buildup Index (BUI), and FWI, derived from prior research. Figure 5 illustrates the FWI process

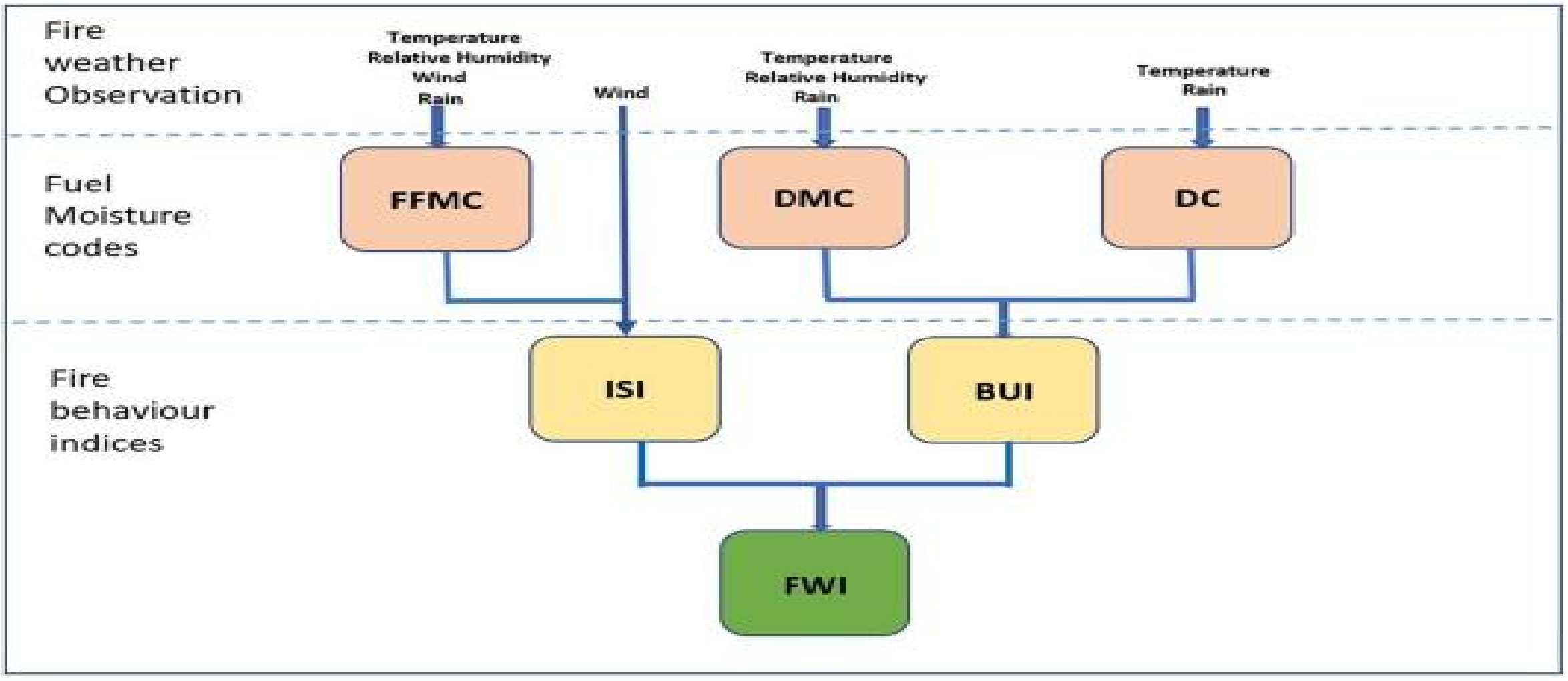

through a detailed flow chart.

Figure 5: Flowchart of the Fire Weather Index [25]

Figure 6 provides detailed information on how rules are applied to streaming data. DC rules are applied to batches of size 20, yielding results based on specific parameters. These results, categorized as "Mop up Needs," encompass various parameters as depicted in figure 6.

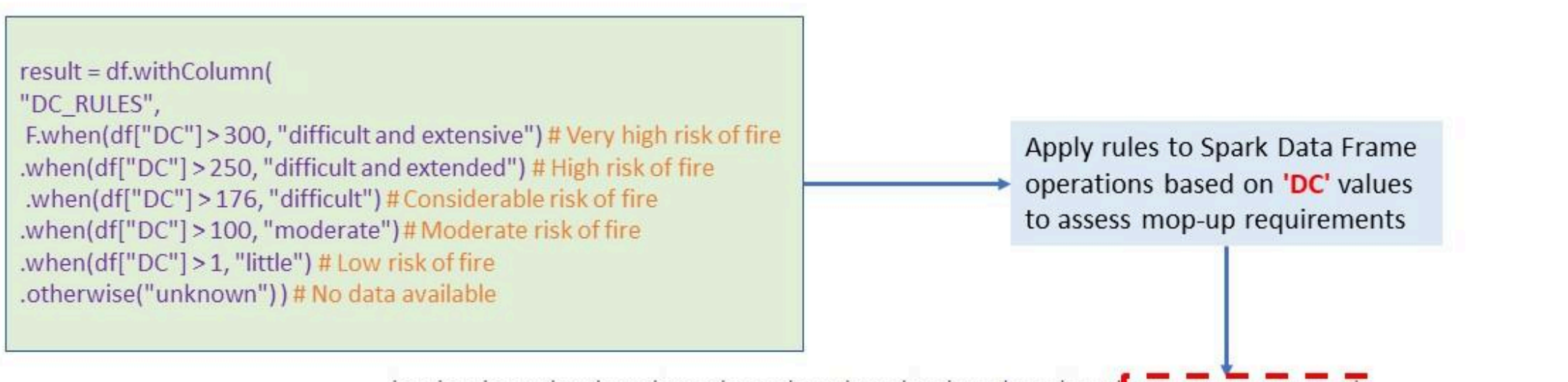

| X | Y | month | day | FFMC | DMC | DC | ISI | temp | RH | wind | rain | area | MopupNeeds |
|---|---|---|---|---|---|---|---|---|---|---|---|---|---|
| 7 | 5 | mar | fri | 86.2 | 26.2 | 94.3 | 5.1 | 8.2 | 51 | 6.7 | 0.0 | 0.0 | little |
| 7 | 4 | oct | tue | 90.6 | 35.4 | 669.1 | 6.7 | 18.0 | 33 | 0.9 | 0.0 | 0.0 | difficult and ext... |
| 7 | 4 | oct | sat | 90.6 | 43.7 | 686.9 | 6.7 | 14.6 | 33 | 1.3 | 0.0 | 0.0 | difficult and ext... |
| 8 | 6 | mar | fri | 91.7 | 33.3 | 77.5 | 9.0 | 8.3 | 97 | 4.0 | 0.2 | 0.0 | little |
| 8 | 6 | mar | sun | 89.3 | 51.3 | 102.2 | 9.6 | 11.4 | 99 | 1.8 | 0.0 | 0.0 | moderate |
| 8 | 6 | aug | sun | 92.3 | 85.3 | 488.0 | 14.7 | 22.2 | 29 | 5.4 | 0.0 | 0.0 | difficult and ext... |
| 8 | 6 | aug | mon | 92.3 | 88.9 | 495.6 | 8.5 | 24.1 | 27 | 3.1 | 0.0 | 0.0 | difficult and ext... |
| 8 | 6 | aug | mon | 91.5 | 145.4 | 608.2 | 10.7 | 8.0 | 86 | 2.2 | 0.0 | 0.0 | difficult and ext... |
| 8 | 6 | sep | tue | 91.0 | 129.5 | 692.6 | 7.0 | 13.1 | 63 | 5.4 | 0.0 | 0.0 | difficult and ext... |
| 7 | 5 | sep | sat | 92.5 | 88.0 | 698.6 | 7.1 | 22.8 | 40 | 4.0 | 0.0 | 0.0 | difficult and ext... |
| 7 | 5 | sep | sat | 92.5 | 88.0 | 698.6 | 7.1 | 17.8 | 51 | 7.2 | 0.0 | 0.0 | difficult and ext... |
| 7 | 5 | sep | sat | 92.8 | 73.2 | 713.0 | 22.6 | 19.3 | 38 | 4.0 | 0.0 | 0.0 | difficult and ext... |
| 6 | 5 | aug | fri | 63.5 | 70.8 | 665.3 | 0.8 | 17.0 | 72 | 6.7 | 0.0 | 0.0 | difficult and ext... |
| 6 | 5 | sep | mon | 90.9 | 126.5 | 686.5 | 7.0 | 21.3 | 42 | 2.2 | 0.0 | 0.0 | difficult and ext... |
| 6 | 5 | sep | wed | 92.9 | 133.3 | 699.6 | 9.2 | 26.4 | 21 | 4.5 | 0.0 | 0.0 | difficult and ext... |
| 6 | 5 | sep | fri | 93.3 | 141.2 | 713.9 | 13.9 | 22.9 | 44 | 5.4 | 0.0 | 0.0 | difficult and ext... |
| 5 | 5 | mar | sat | 91.7 | 35.8 | 80.8 | 7.8 | 15.1 | 27 | 5.4 | 0.0 | 0.0 | little |
| 8 | 5 | oct | mon | 84.9 | 32.8 | 664.2 | 3.0 | 16.7 | 47 | 4.9 | 0.0 | 0.0 | difficult and ext... |
| 6 | 4 | mar | wed | 89.2 | 27.9 | 70.8 | 6.3 | 15.9 | 35 | 4.0 | 0.0 | 0.0 | little |
| 6 | 4 | apr | sat | 86.3 | 27.4 | 97.1 | 5.1 | 9.3 | 44 | 4.5 | 0.0 | 0.0 | little |

Figure 6: Rules applied on DC values on stream data and tell about mop-up Needs

DC elucidates the necessity of mop-up operations, highlighting the degree of controllable fire risk and facilitating appropriate action. The forest ontology contains all of the forest's location details, making it easy to take action in areas with different types of forests, whether dense or less dense. We generate other FFMC, ISI, BUI, DMC, and FWI alerts based on stream data in the same manner.

### 3.4.2 Implement an alert system that utilizes LLMs for processing queries based on Spark alerts

The research employs advanced techniques to manage and interpret forest fire data, subsequently transforming it into a refined text format. A thorough preprocessing phase initiates the process, cleansing the data of discrepancies and noise to ensure reliability and quality. Advanced text normalization methods normalize the clean data to achieve uniformity across the dataset, preparing it for further analytical processes.

We converted the normalized text into document embeddings using the state-of-the-art, all-Mini LM-L6-v2 model. These embeddings are critical because they capture the text's deep semantic meanings, essential for accurately retrieving information based on the content's relevance and similarity. The embedding process preserves and makes meaningfully retrievable the useless and critical information within the text. We build an indexing system using FAISS (Facebook AI Similarity Search) to retrieve these semantically rich document embeddings, providing a robust framework for indexing and quickly retrieving the documents. This indexing is critical for supporting the rapid access required in the querying phase, enhancing the retrieval process's efficiency.

The question-answering setup is the core of the proposed system, which integrates the Intel/dynamic_tinybert model within a retrieval-based framework. For each posed query, this setup precisely configures a retriever to identify and pull the two most relevant documents from the index. As a result, a retrieval QA model uses these documents to generate accurate answers. A conversation buffer memory further refines this model by maintaining context over successive interactions, thereby ensuring the system's responses are consistent and relevant.

Additionally, the system's real-time capabilities allow it to handle immediate queries about environmental conditions, sensor readings, or alarm activations, delivering prompt and accurate responses. Integration with Apache Spark enhances this capability by distributing processing tasks across multiple nodes, ensuring that the system can scale effectively to handle large datasets and simultaneous queries without a loss in performance.

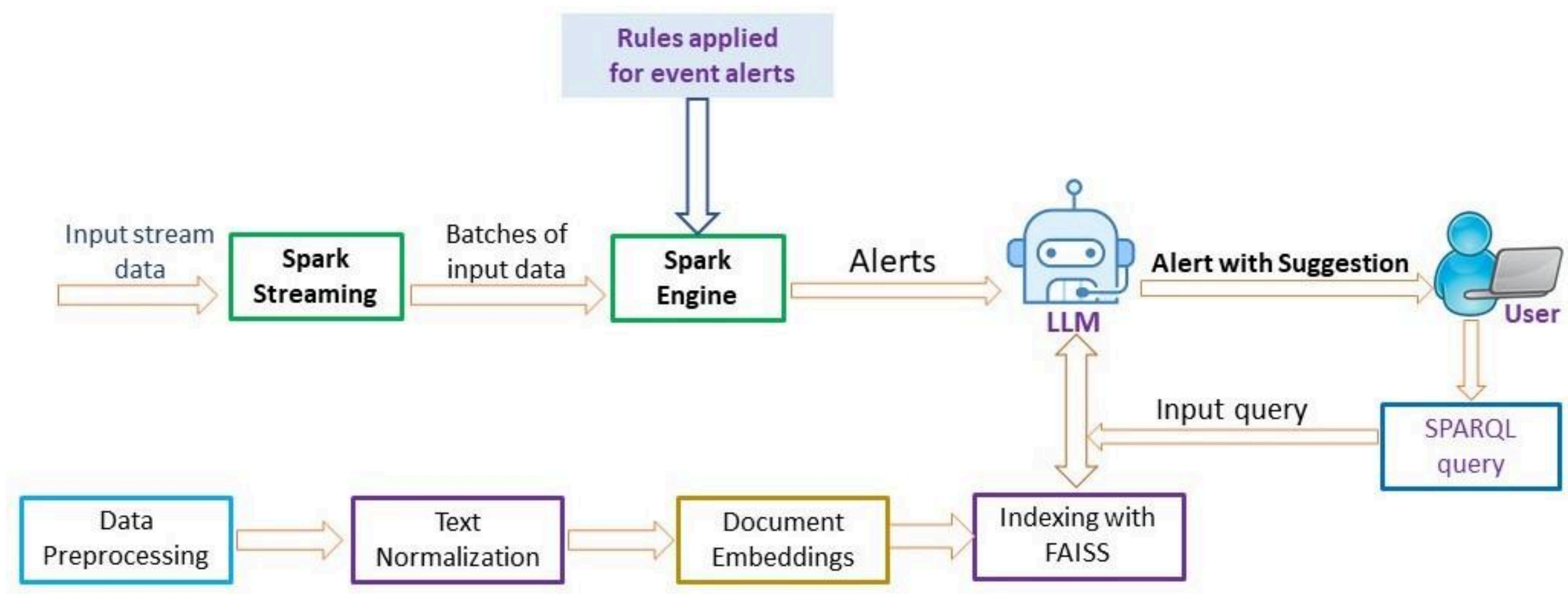

Figure 7: Integration of spark framework with LLMs

The deployment of pre-trained LLMs alongside Apache Spark boosts the system's speed and efficiency and its adaptability and scalability. Figure 7 shows that this makes sure that as fire safety rules change along with improvements in sensor technology and changes in the environment, the system can adapt and continue to provide reliable, cost-effective fire alarm trigger detection, better protecting lives and property.

### 3.5 SPARQL query-based LLMs alert and ontology alert generation

When the same SPARQL query is executed, the results from both the LLMs-based alert system and the ontology-based alert system are nearly identical. However, leveraging its advanced learning capabilities, the LLM system frequently provides additional insights and suggestions, which are shown in Table 6. Subsequently, the ontology system evaluates these results from the LLMs, employing its extensive and detailed knowledge base. This base includes critical information such as location specifics, temperature readings, and data pertaining to sensor deployment areas, thereby ensuring that the LLMs suggestions are both relevant and accurate.

Table 6: SPARQL query and corresponding result.

| | **Query** | **Result** |
|---|---|---|
| SPARQL query ontology | SELECT ?region ?temperature ?humidity<br>WHERE {<br>?region rdf:type ex:ForestArea .<br>?region ex:hasTemperature?temperature .<br>?region ex:hasHumidity ?humidity .<br>FILTER (?temperature > 30 && ?humidity < 30) | -----------------------------------------------<br>**\| region \| temperature \| humidity \|**<br>-----------------------------------------------<br>\| "Pine Valley" \| 35 \| 25 \|<br>\| "Oak Ridge" \| 34 \| 20 \|<br>\| "Maple Hill" \| 32 \| 29 \| |
| | } | |
| SQL query on LLMs | SELECT region, temperature, humidity FROM ForestAreas<br>WHERE temperature > 30 AND humidity < 30; | **Predictive Insight:** Based on the current dry spell and wind conditions, South Forest, although currently at 28 degrees and 32% humidity, will likely exceed high-risk thresholds within the next 48 hours.<br><br>**Precautionary measures are recommended:**<br>Resource Allocation Suggestion: Due to the severe risk parameters observed, immediate fire watch alerts and additional surveillance drones are suggested for North Forest. |

Figure 8 shows how we ask questions to both the ontology and LLMs. It demonstrates how LLMs find information using a system called FAISS. This helps us understand how both methods work together to get the necessary information.

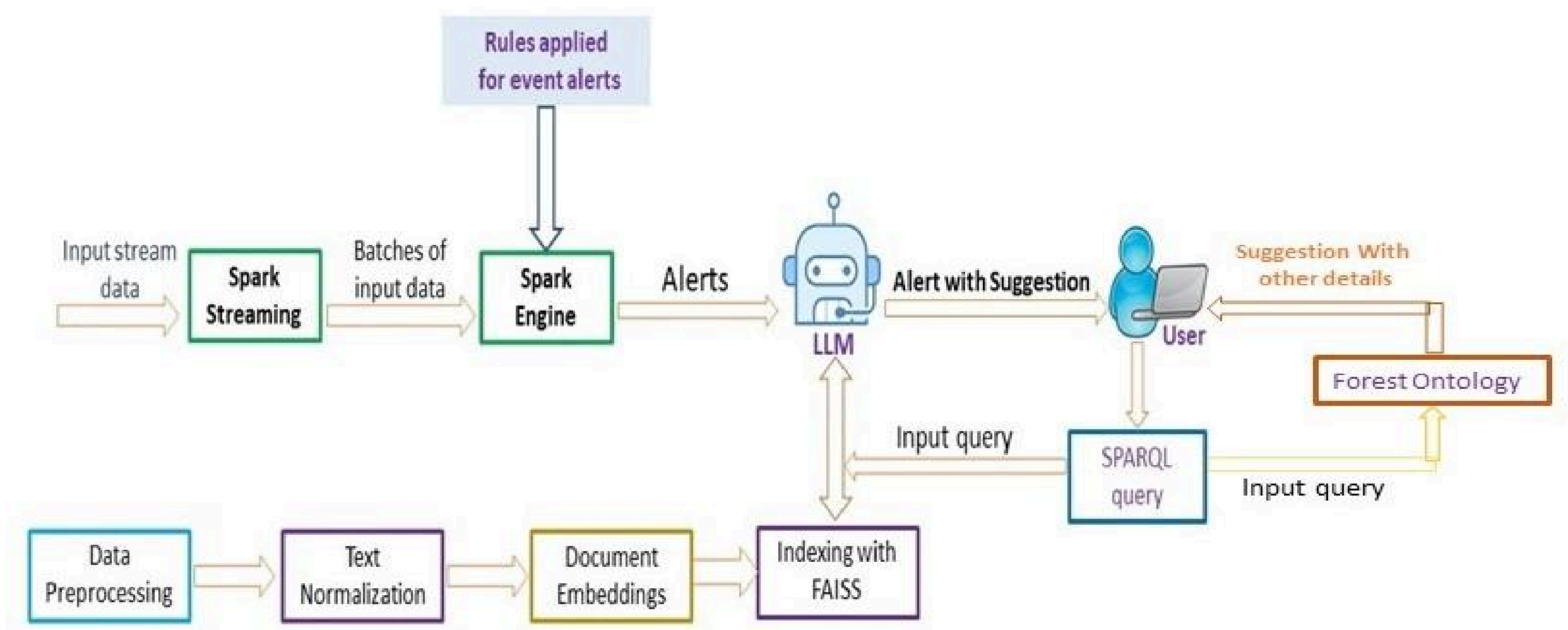

Figure 8: Integration of spark framework with LLMs and Ontology

## 4. Result and Discussion

This research aims to develop a DSS to help prevent and manage forest fires. In our previous work, we used sensor data from Montesinho Natural Park for its creation. We utilize four types of weather sensors: precipitation, wind speed, humidity, and temperature. Additionally, we incorporate survey reports to supplement the data for our model. Our ontology specifies sensor locations based on these reports. Using SWRL rules, our ontology provides precise recommendations when processed with the reasoner in the Protege tool. We have integrated 43 rules into our model, which was developed with the help of the SSN ontology. We use SPARQL queries to obtain comprehensive results, executing 97 queries on this model and producing accurate outcomes [25].

To extend this work, we enhanced the ontology based on forest management system documents from countries like Australia, Canada, and India, making our forest ontology more precise and comprehensive. We've also introduced Spark streaming-based alert methods and LLMs to automate the system, reducing human intervention. We consulted numerous forest documentation sources for LLMs development, which provide results and offer precautionary suggestions tailored to specific situations. We've added over 200 SWRL rules to the ontology, which are also used in Spark's streaming alert detection. These alerts are sent to the LLMs, which provide precautionary suggestions based on the situation, enhancing accuracy. These alerts are applied in the ontology, and results are obtained, informing actions taken by the forest department. In further sections, we evaluate our models.

### 4.1 Evaluation of the forest ontology based on metric counts

Based on the ontology metrics count shown in Table 7, we define the inheritance richness, relationship richness, class relation ratio, axiom, class ratio, and inheritance richness, as detailed in Table 8.

Table 7: Metrics available in the Forest ontology.

| | |
|---|---|
| Axioms | 5897 |
| Logical axioms count | 2129 |
| Class count | 1007 |
| Data property count | 198 |
| Object property count | 454 |
| Annotation Property count | 53 |
| Object Property Range | 281 |
| Object Property Domain | 483 |
| Individual count | 587 |
| Total individuals count | 407 |

**Schema metrics:** The ontology can also be described using a 5-tuple model, represented as O = <C, Dr, Sc, Re, Ind>, where C stands for classes, Dr denotes data properties (attributes), Sc indicates subclasses, Re represents relations between classes, and Ind signifies individuals. Metrics can be evaluated based on Attribute Richness, Relationship Richness, Class Richness, and Average Population.

**RR** measures the depth of connections between concepts within an ontology. It is determined using Equation 1:

$$RR = |Prop| / |Sub\,class| + |Prop| \quad \ldots\ldots\ldots\ldots\ldots (1)$$

where |Prop| represents the total number of properties, encompassing both attribute data and object characteristics (class relationships).

**AR** is calculated by averaging the number of attributes over the entire class, as shown in Equation 2:

AR = |Attribute| / |Class| ………………….. (2)

where |attribute| represents the total number of data attributes.

**CR** indicates the extent of real-world knowledge represented by the ontology. It is calculated using Equation 3, by dividing the number of classes with instances by the total number of classes:

CR = | Class with_instance | / |Class| ……………….. (3)

**AP** determines the average number of individuals in each class, expressed in Equation 4:

AP = | Individual | / |Class| …………………….(4)

The computed values of different evaluation metrics for the newly designed ontology are mentioned in table 8.

Table 8: Measurements and values associated with ontology

| Schema Metrics | Value | Value [25] |
|---|---|---|
| Relationship richness | 0.8703 | 0.8093 |
| Class/relation ratio | 0.9781 | 0.7948 |
| Inheritance richness | 0.8961 | 0.8587 |
| Axiom/class ratio | 67.2622 | 42.1742 |
| Attribute richness | 0.9019 | 0.8289 |

We also assess the quality score of the ontology based on the represented knowledge, which reflects the quantity and relationships of the domain knowledge. The score for the ontology model (om) can be calculated using the following equations (5).

Score om = [(| Rel | * |Class| × 100) + (|Subclass|+ |Rel|) * |Prop|) / (|Subclass| + |Rel|) * |Class|] …………………………..(5)

An ontology score can also be calculated based on the effectiveness of knowledge base (kb) extraction, following Equation (6).

Score kb = [(Number of Classes × 100) + (|Individual|)/ |Class|] ………………….(6)

Table 9 presents the ontology score calculated using the formulas for scores om and kb.

Table 9: Forest ontology score.

| Evaluation parameter | Score | Score [25] |
|---|---|---|
| Score om | 96.04 | 89.05 |
| Score kb | 95.02 | 92.03 |

**4.2 LLMs based evaluation**

To measure the performance of LLMs, we first extract relevant context from the FAISS database based on the input query. The retrieved context is ranked using cosine similarity, which measures the relevance of the context to the query. Cosine similarity values range from -1 to +1, where values closer to +1 indicate higher relevance. After retrieving the context, it is fed into the LLMs, which generate a response based on this information. We then evaluate the LLM's output by calculating Precision, Recall, and F-measure, comparing the generated response against a predefined test response. Precision indicates the proportion of relevant information in the response, Recall measures how much relevant information from the context was captured, and the F-measure provides a balance between Precision and Recall. High values for these metrics suggest that the LLM's response is both accurate and contextually relevant, validating its performance.

Table 10: Query-based LLMs score

| Query | Cosine Similarity Score(FAISS context) | Score of response | | |
|---|---|---|---|---|
| | | Precision | Recall | F-measure |
| If Ignition Potential is extremely easy and DMC Rules are difficult, what actions should be taken to reduce forest fire? | 0.82604045 | 0.723 | 0.685 | 0.7025 |
| If Ignition Potential is extremely easy, DMC Rules is difficult and extensive, DC Rules is difficult and extensive, Rate of Spread is fast and raining is unknown and wind speed is unknown then what measures should be taken? | 0.8750995 | 0.680 | 0.750 | 0.7118 |
| If Ignition Potential is moderately easy, DMC Rules is difficult and extensive, DC Rules is difficult and extensive, the Rate of Spread is slow, while Raining is unknown and Wind Speed is unknown, then what measures should be taken? | 0.8609843 | 0.750 | 0.779 | 0.7638 |
| The Ignition Potential is moderately easy, DMC Rules is difficult and extensive,DC Rules is difficult and extensive while the Rate of Spread is slow, the raining status and speed of wind is unknown. What measures should be taken? | 0.8217561 | 0.867 | 0.654 | 0.7465 |
| The Ignition Potential is moderately easy,DMC Rules is difficult and extensive and DC Rules is difficult and extensive while the Rate of Spread is slow the raining status and spead of wind is unknown what measures should be taken? | 0.88260404 | 0.786 | 0.890 | 0.8329 |

In table 10 the first query was shown in figure 9 and its results with proper suggestion and precaution measures.

```
question = '''If Ignition Potential is extremely easy, DMC Rules is difficult and extensive,
 DC Rules is difficult and extensive, Rate of Spread is fast and raining is unknown and wind speed
 is unknown then what measures should be taken?'''

response = llm_chain.invoke(f"{question} answer by underatanding this context {context}")

response['text']

' In such a situation where the fire potential is low but the area is vast and extensive with various challenges, the following measures could be taken to help in containing the fire and preventing its spread:\n\n1. Develop an Incident Action Plan (IAP): Create an incident action plan that outlines the tactics and techniques for fighting the fire. The IAP should consist of a detailed description of the fire's location, size, and behavior, as well as the resources available to fight it.\n2. Establish Communication Channels: Set up reliable communication channels between the incident command, firefighters, and other stakeholders. This will help to disseminate vital information and ensure that everyone is on the same page.\n3. Assign Tasks: Assign tasks to firefighters based on their strengths, skills, and experience. This ensures that each team member has a specific role in containing the fire.\n4. Use Firefighting Techniques: Employ tried-and-true firefighting methods such as direct atta...'
```

Figure 9. LLMs based query result

Table 11 presents the outcomes of LLM, which do not incorporate external documentation guidelines; it solely utilizes the dataset and the Monterio Natural Park survey report, both of which were employed in the development of the ontology in prior studies. Derived from the development of these LLMs, we obtain the results of cosine similarity and the LLMs' answers in the form of true or false after querying the LLMs.

Table 11: Query-based cosine similarity and LLMs response

| Query | Cosine Similarity | LLMs answer |
|---|---|---|
| If X: 8, Y: 6, Month: aug, Day: mon, FFMC: 92.3, DMC: 88.9, DC: 495.6, ISI: 8.5, Temperature: 24.1, RH: 27, Wind: 3.1, Rain: 0.0, Area: 0.0, Ignition Potential: extremely easy, DMC Rules: difficult and extensive, DC Rules: difficult and extensive, Rate of Spread: fast, Start Raining: unknown, Wind Speed: unknown, then should Fire Trigger should be sent? | 0.879 | True |
| If X: 1, Y: 4, Month: aug, Day: sat, FFMC: 94.4, DMC: 146.0, DC: 614.7, ISI: 11.3, Temperature: 25.6, RH: 42, Wind: 4.0, Rain: 0.0, Area: 0.0, Ignition Potential: extremely easy, DMC Rules: difficult and extensive, DC Rules: difficult and extensive, Rate of Spread: fast, Start Raining: unknown, Wind Speed: unknown, then should Fire Trigger be sent? | 0.882 | True |
| If X: 7, Y: 4, Month: aug, Day: sun, FFMC: 81.6, DMC: 56.7, DC: 665.6, ISI: 1.9, Temperature: 21.2, RH: 70, Wind: 6.7, Rain: 0.0, Area: 11.16, Ignition Potential: moderately easy, DMC Rules: difficult and extensive, DC Rules: difficult and extensive, Rate of Spread: slow, Start Raining: unknown, Wind Speed: unknown, then should Fire Trigger should be sent? | 0.848 | True |
| If X: 2, Y: 4, Month: aug, Day: sun, FFMC: 81.6, DMC: 56.7, DC: 665.6, ISI: 1.9, Temperature: 21.9, RH: 71, Wind: 5.8, Rain: 0.0, Area: 54.29, Ignition Potential: moderately easy, DMC Rules: difficult and extensive, DC Rules: difficult and extensive, Rate of Spread: slow, Start Raining: unknown, Wind Speed: unknown, then should Fire Trigger be sent? | 0.883 | True |
| If X: 4, Y: 3, Month: aug, Day: sun, FFMC: 81.6, DMC: 56.7, DC: 665.6, ISI: 1.9, Temperature: 27.8, RH: 32, Wind: 2.7, Rain: 0.0, Area: 6.44, Ignition Potential: moderately easy, DMC Rules: difficult and extensive, DC Rules: difficult and extensive, Rate of Spread: slow, Start Raining: unknown, Wind Speed: unknown, then should Fire Trigger should be sent? | 0.873 | True |

These are sample queries used to test the embeddings, and the returned context was evaluated based on cosine similarity. The LLMs' responses were tested against a test response and have proven to be accurate. The embeddings were stored based on various features such as X and Y coordinates, month, day of the week, FFMC, DMC, DC, ISI, temperature, RH, wind speed, rainfall status, and ignition area. Ground truth rules were applied to these values, indicating whether a fire alarm trigger should be sent or not. The retriever then shifts through the embedding database and retrieves data relevant to the query based on cosine similarity. Subsequently, the model answers based on the received context. The results demonstrate the high accuracy of the retrieved context from the embeddings, facilitating effective decision-making by the LLMs.

### 4.3 SPARQL query performance

Figure 10 depicts the query executed via the LLMs ontology, representing it as an individual component in the proposed work. However, similar queries were executed through combinations of the LLMs ontology and Spark. It's observed that the execution time for Q3 and Q5 varies significantly, with combined models taking more time for the same queries compared to individual models due to complex query. Table 12 illustrates how our model surpasses existing works.

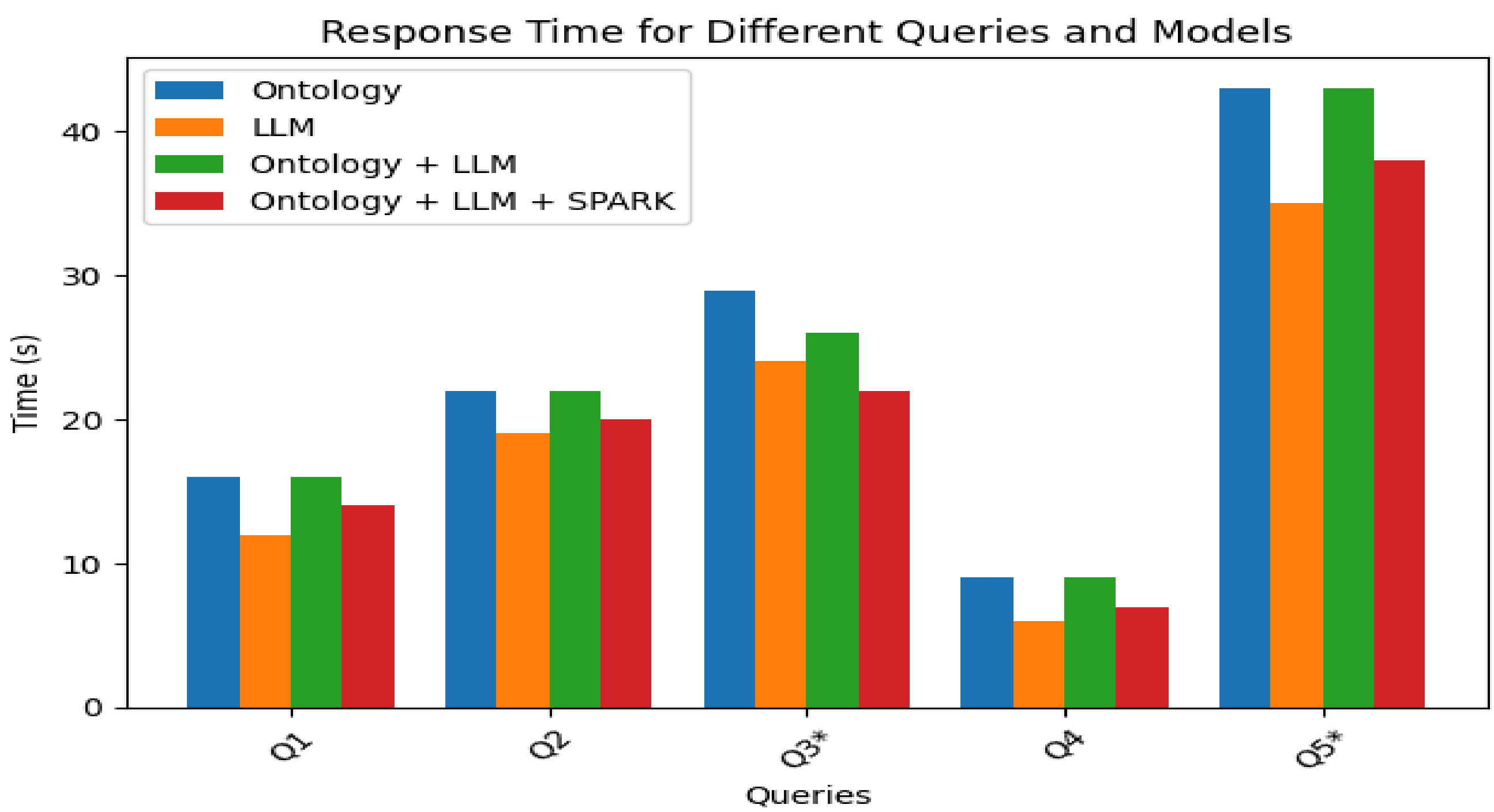

Figure 10: SPARQL query performance analysis

### 4.4 Comparison with existing literature work.

In this section, we compare the results with existing methods, as shown in Table 12.

Table 12: Comparison with existing work

| Reference | Objective | Methodology | Technology used | Result |
|---|---|---|---|---|
| Chandra et al.[25] | Forest fire management using semantic sensor technology. | SSN ontology and SWRL rules based DSS system. | OWL language for ontology and SWRl rules development | Ontology metrics based evaluation, Query based evaluation. |
| Masa et al.[26] | A framework based on ontology for representing and interlinking wildfire events, utilized for decision support in forest fire management. | ONTO-SAFE ontology. | Ontological web language and semantic reasoning. | Incorporating expert knowledge in the form of constraints and rules to identify patterns and situations within domain knowledge, aiding end-users in advanced decision-making. |
| Gao et al.[27] | Integration of semantic reasoning and ontology to design semantic fire weather index. | Rule based inference algorithm and Inverse Distance Weighting (IDW). | Wireless sensor network and Semantic reasoning. | Queries used to evaluate the models efficiently in terms of accuracy and precision. |
| Ge et al.[28] | Forest fire analysis using the semantic rules of the knowledge graph. | Spatio-temporal knowledge graph. | Machine learning and rule based reasoning. | Real time fire prediction using precision, recall and F1 score. |
| **Proposed Model** | Forest Fire Management. | Forest ontology, SWRL rules, Spark streaming based alert, and LLMs based precautionary suggestion. | OWL web language, LLMs, Spark. | Cosine score, Precision,recall, F1 score, Ontology based metrics evaluation. Attribute Richness, Class Richness, Relationship Richness, and Average Population. |

## Conclusion and Future Work

This paper utilizes ontology, LLMs, and Spark streaming to detect forest fires in the forest fire management system. Previous research has primarily focused on ontology and SWRL rule-based forest fire management, which require significant human intervention. To get around this problem, this study uses Spark streaming to process data in real time and create rules-based alerts. Additionally, it utilizes LLMs and ontology to generate safety recommendations and determine appropriate actions. This integration minimizes human intervention and enables instant action orders from forest officers to ground action teams. To evaluate the performance of the model, we assessed the ontology based on metrics such as schema, query response time, LLMs response score, and cosine similarity. Various scores demonstrate that the developed ontology encompasses a broader range of forest management knowledge compared to previous versions. This comprehensive model yields more accurate results. However, a limitation of this approach is the need for knowledge amendments when deploying the model in different locations. It's essential to include area-specific knowledge for the model to adapt to local conditions.

For future work, we aim to enhance the model's efficiency and reliability by incorporating additional rules and integrating machine learning techniques. This ongoing development will further optimize the system's performance and adaptability to varying forest fire scenarios.

## Declaration

### Competing interests

The authors declare no competing interests relevant to the content of this work.

### Author’s contribution statement

All authors contributed equally to this work.

**Data availability and access**

The study utilized freely accessible data obtained from a website. We extend our thanks to the authors and collaborators for providing the original data.

**Funding and Acknowledgment**

The authors express their gratitude to the Indian Institute of Information Technology, Allahabad, for providing the necessary materials needed to conduct this study.

**Conflicts of interests**

All authors declare that they have no conflicts of interest in the presented work.